\documentclass[conference,a4paper]{IEEEtran}
  \IEEEoverridecommandlockouts
  \usepackage{cite}
  \usepackage{amsmath,amssymb,amsfonts}
  \usepackage{graphicx}
  \usepackage{textcomp}
  \usepackage{multirow}
  \usepackage{xcolor}
  \usepackage{booktabs}
  
\begin{document}

  \title{Learned Radius Estimation for UDF-Based \\Point Cloud Reconstruction}
  
  \author{
    \IEEEauthorblockN{Eito Ogawa}
    \IEEEauthorblockA{\textit{Graduate School of FSE} \\
    \textit{Waseda University}\\
    Tokyo, Japan \\   
    ogawaeito@fuji.waseda.jp}
    \and
    \IEEEauthorblockN{Hiroshi Watanabe}
    \IEEEauthorblockA{\textit{Graduate School of FSE} \\
    \textit{Waseda University}\\
    Tokyo, Japan \\
    hiroshi.watanabe@waseda.jp}
  }

  \maketitle

  \begin{abstract}
Surface reconstruction from point clouds is important for consumer-grade
3D capture, including AR/VR and indoor scanning. Local-patch Unsigned
Distance Field (UDF) methods are lightweight and generalizable, but
their accuracy depends on the support radius, traditionally fixed or
selected by a one-dimensional curvature heuristic that cannot capture
heterogeneous local geometry.

We propose a learned per-query radius selector that predicts a
continuous support radius and plugs into a frozen LoSF-UDF backbone.
The selector is trained using off-grid target radii obtained by
parabolic interpolation of cached UDF error curves. Experiments show
improved fine-scale reconstruction accuracy, with ScanNet F1@0.005
improving from 0.645 to 0.691 over a curvature-based baseline.

  \end{abstract}
  
  
  \section{Introduction}
  With the widespread use of smartphones, AR/VR devices, and home robots,
  efficient 3D reconstruction for consumer-grade capture has become
  increasingly important. Point clouds captured by such consumer devices
  typically exhibit non-uniform sampling density and local sparsity,
  requiring reconstruction methods that robustly handle both closed and open surfaces. Unsigned Distance Fields (UDFs)~\cite{ndf} avoid
  inside/outside classification, and LoSF-UDF~\cite{losfudf} estimates
  UDF values locally from point-cloud patches in a lightweight manner.
  Our previous method, GeoLA~\cite{ievc}, adapts the local support radius
  via a curvature heuristic. However, a scalar curvature descriptor
  cannot capture heterogeneous local geometry.
  
  In this paper, we propose a learned per-query radius selector that
replaces this heuristic with a data-driven mapping. The selector is
trained using interpolated target radii derived from local parabolic
interpolation of cached UDF error curves, providing off-grid supervision
beyond the discrete candidate set. It plugs into the frozen LoSF-UDF
backbone without retraining while keeping the overall pipeline compact.

  \begin{figure}[t]  
      \centering
      \includegraphics[width=\columnwidth]{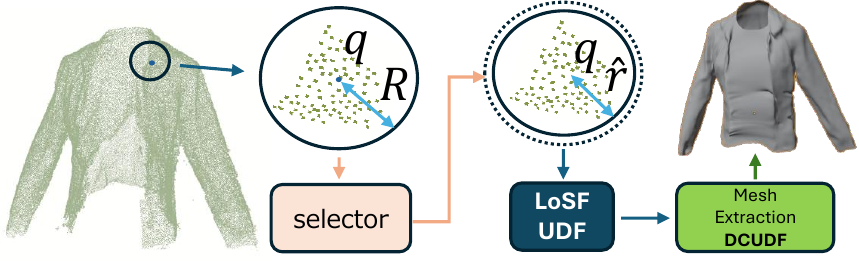}
      \caption{Overview of the proposed framework. For each query point, a parent patch
with radius $R$ is extracted, from which the selector predicts an adaptive
support radius $\hat{r}$. The selected patch with radius $\hat{r}$ is fed
into the frozen LoSF-UDF backbone to estimate the UDF value, which is then
used for mesh extraction with DCUDF~\cite{dcudf}.}
      \label{fig:reconstruction_pipeline}
  \end{figure}

\begin{table*}[t]
\centering
\caption{Quantitative results}
\label{tab:main}
\footnotesize
\setlength{\tabcolsep}{3pt}
\renewcommand{\arraystretch}{0.95}

\begin{tabular*}{\textwidth}{@{\extracolsep{\fill}}l|cccc|cccc|cccc@{}}
\hline
& \multicolumn{4}{c|}{Cars~\cite{shapenet}}
& \multicolumn{4}{c|}{Fashion~\cite{deepfashion3d}}
& \multicolumn{4}{c}{ScanNet~\cite{scannet}} \tabularnewline
Method
& CD$\downarrow$ & F1$^{0.005}\uparrow$ & F1$^{0.01}\uparrow$ & NC$\uparrow$
& CD$\downarrow$ & F1$^{0.005}\uparrow$ & F1$^{0.01}\uparrow$ & NC$\uparrow$
& CD$\downarrow$ & F1$^{0.005}\uparrow$ & F1$^{0.01}\uparrow$ & NC$\uparrow$ \tabularnewline
\hline\hline
LoSF-UDF~\cite{losfudf}
& 0.794 & 0.506 & 0.865 & 0.594
& 0.442 & 0.646 & 0.941 & \textbf{0.951}
& 0.715 & 0.651 & \textbf{0.948} & \textbf{0.876} \tabularnewline
GeoLA~\cite{ievc}
& 0.732 & 0.571 & 0.893 & 0.808
& 0.422 & 0.657 & 0.945 & 0.946
& 0.795 & 0.645 & 0.899 & 0.824 \tabularnewline
Ours
& \textbf{0.705} & \textbf{0.581} & \textbf{0.896} & \textbf{0.822}
& \textbf{0.404} & \textbf{0.681} & \textbf{0.946} & 0.942
& \textbf{0.692} & \textbf{0.691} & 0.945 & 0.855 \tabularnewline
\hline
\end{tabular*}
\end{table*}

  \section{Related Work}

  Implicit neural representations such as signed distance and occupancy
  fields require inside/outside classification, limiting them to closed
  surfaces. UDFs avoid this by representing the unsigned distance to the
  nearest surface, supporting both closed and open surfaces. LoSF-UDF
  \cite{losfudf} estimates the UDF locally from a neighborhood patch
  around each query point, rather than compressing the point cloud
  into a global latent representation. This local design yields a lightweight network that generalizes across shapes. However, the patch radius is
  fixed across the dataset: an overly large radius causes multi-surface
  contamination at fine details, while an overly small one provides
  insufficient local context for UDF estimation.

  \section{Proposed Method}
  Our framework consists of (i) an offline procedure that derives
interpolated target radii from cached UDF error curves, and (ii) a
  lightweight selector that regresses the per-query radius from the
  parent patch and plugs into the frozen LoSF-UDF backbone without
  retraining.
  
  \subsection{Target Radius Estimation via Local Parabolic Interpolation}

  For each query $q$, we construct $K$ local patches at candidate radii
  $\{r_1, \ldots, r_K\}$ and pass them through the frozen LoSF-UDF
  backbone, obtaining a per-query error vector $\mathbf{e} \in
  \mathbb{R}^{K}$ against the ground truth. From the discrete argmin
  $k^{*} = \arg\min_{k} e_{k}$, we fit a parabola through
  $(e_{k^{*}-1}, e_{k^{*}}, e_{k^{*}+1})$ and take its vertex as the interpolated target radius $r^{*}$
  \begin{equation}
    \delta = \frac{1}{2} \cdot
             \frac{e_{k^{*}-1} - e_{k^{*}+1}}
                  {e_{k^{*}-1} - 2 e_{k^{*}} + e_{k^{*}+1}},
    \quad
    r^{*} = r_{k^{*}} + \delta \cdot \Delta r,
  \end{equation}
  where $\Delta r$ is the candidate spacing. On the boundary or in
  non-convex neighborhoods, we fall back to $r_{k^{*}}$. This provides an off-grid regression target instead of a discrete candidate label. These cached target radii are used only during training; no test-time ground truth
  is required.

  \subsection{Selector Architecture and Loss}

  The parent-patch points $P \in \mathbb{R}^{N \times 3}$ are encoded by
  a masked ResNet-PointNet into a 256-dimensional feature
  $\boldsymbol{\phi}$, concatenated with the point count $\log(1+N)$ and
  a radial density histogram $\mathbf{h} \in \mathbb{R}^{6}$. A
  three-layer MLP with sigmoid output yields a continuous radius ratio
  $\rho \in [\rho_{\min}, \rho_{\max}]$, giving the final radius
  $\hat{r} = \rho \cdot R_{\text{parent}}$. We train with a
  confidence-weighted normalized L1 loss:
  \begin{equation}
    \mathcal{L} = w \cdot
                  \frac{|\hat{r} - r^{*}|}{r_{\max} - r_{\min}},
  \end{equation}
  where $w = \max \mathbf{e} - \min \mathbf{e}$ emphasizes samples whose
  UDF error is sensitive to the radius choice. At inference, a single
  selector forward pass yields $\hat{r}$, independent of $K$ used during
  training.

  \section{Experiments}
We evaluate surface reconstruction accuracy on ScanNet~\cite{scannet},
ShapeNet-Cars~\cite{shapenet}, and DeepFashion3D~\cite{deepfashion3d},
using 100 randomly sampled test shapes from each dataset.
 For selector training, we use 40 cars from
  ShapeNet-Cars and 40 garments from DeepFashion3D, disjoint from the
  test shapes. ScanNet is held out entirely from training, providing a
  cross-domain evaluation of generalization to real indoor scans.
We use $K=7$ candidate radii in $\{0.012, 0.014, \ldots, 0.024\}$ with parent radius $R_{\mathrm{parent}}=0.024$, corresponding to $\rho \in [0.5, 1.0]$, and adopt the released pretrained LoSF-UDF~\cite{losfudf} backbone, kept frozen during selector training. Final meshes are extracted from the predicted UDF field using DCUDF~\cite{dcudf}. We report bidirectional L1 Chamfer Distance (CD, $\times 10^2$), F-score at thresholds $0.005$ and $0.01$,
  and Normal Consistency (NC).

  Table~\ref{tab:main} compares our method against the curvature-based
  GeoLA~\cite{ievc} and the fixed-radius LoSF-UDF~\cite{losfudf}.
  Our method achieves the best CD and F1@0.005 across all three
  datasets, including the held-out ScanNet, demonstrating that the
  learned selector generalizes from only 80 CAD/garment training shapes
  to real indoor scans. Over GeoLA on ScanNet, F1@0.005 improves from
  0.645 to 0.691 and CD from 0.795 to 0.692; improvements over GeoLA are
  also observed on ShapeNet-Cars and DeepFashion3D. While the fixed-radius LoSF-UDF retains higher F1@0.01 and NC for ScanNet (0.948 vs. 0.945, 0.876 vs. 0.855), the consistent advantage
  of our method on the strict-distance metrics (CD and F1@0.005)
confirms the benefit of learning radius selection
from local patch features rather than from a single curvature scalar.
  
  \begin{figure}[t]
    \centering
    \includegraphics[width=7.5cm]{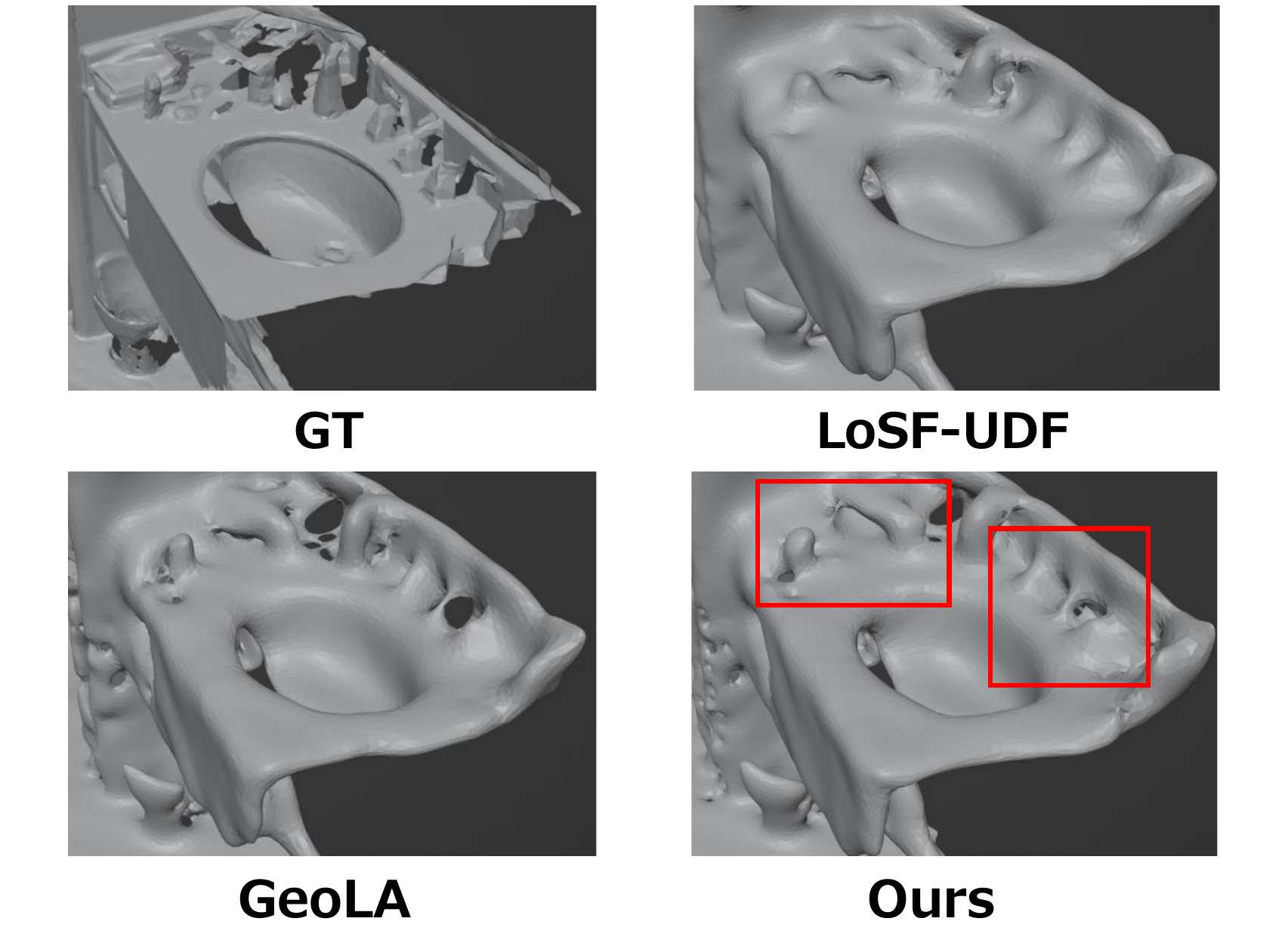}
    \caption{Reconstruction images on ScanNet. From left to right and top to bottom:
GT, LoSF-UDF, GeoLA, and ours. Red boxes highlight regions
where ours preserves fine boundaries and local geometric details.}
    \label{fig:qualitative}
  \end{figure}

  Fig.~\ref{fig:qualitative} shows ScanNet reconstructions: the
  curvature-based GeoLA tends to oversmooth or drop local detail in
  regions with non-uniform sampling such as wall-floor boundaries and
  occlusion edges, whereas the proposed selector preserves the local
  structure through its learned density-aware features.

  \section{Conclusion}
  We proposed a learned per-query radius selector trained using interpolated target radii derived from parabolic interpolation. Trained on only 80 CAD/garment meshes, our method
  generalizes to real indoor scans (ScanNet), outperforming the
  curvature-based prior GeoLA on strict-distance metrics across three
  datasets. The full pipeline (backbone + selector) totals 4.7\,MB of
  fp32 parameters and requires no backbone retraining, making it a
  compact plug-in component for consumer-grade 3D capture. Future work
  includes robustness to noise and missing regions, and extension to
  dynamic scenes.

  \end{document}